\documentclass[conference]{IEEEtran}
\IEEEoverridecommandlockouts

\usepackage{cite}
\usepackage{amsmath,amssymb,amsfonts}
\usepackage{graphicx}
\usepackage{textcomp}
\usepackage{xcolor}
\usepackage{pifont}
\usepackage{booktabs}
\usepackage{multirow}
\usepackage{siunitx}
\usepackage{algorithm}
\usepackage{algorithmicx}
\usepackage{algcompatible}
\usepackage{algpseudocode}
\usepackage{tikz}

\usetikzlibrary{matrix}
\usepackage{pgfplots}
\usepgfplotslibrary{groupplots}
\usepgfplotslibrary{external}
\usepgfplotslibrary{colormaps}
\usepgfplotslibrary{colorbrewer}
\usepackage{pgfplotstable}
\pgfplotsset{compat=1.16}

\usepackage{stfloats}

\newcommand{\cmark}{\textcolor{green!60!black}{\ding{51}}}
\newcommand{\xmark}{\textcolor{red!70!black}{\ding{55}}}

\makeatletter
\def\IEEEauthorrefmark#1{\textsuperscript{#1}}
\makeatother

\begin{document}

\title{Federated Imputation under Heterogeneous Feature Spaces}
\author{

\IEEEauthorblockN{
Imane Hocine\IEEEauthorrefmark{*}\IEEEauthorrefmark{1},
Chaimaa Medjadji\IEEEauthorrefmark{*}\IEEEauthorrefmark{1},
Sylvain Kubler\IEEEauthorrefmark{1}\IEEEauthorrefmark{2},
Grégoire Danoy\IEEEauthorrefmark{1}\IEEEauthorrefmark{2},
Yves Le Traon\IEEEauthorrefmark{1}\IEEEauthorrefmark{2}
}

\IEEEauthorblockA{\IEEEauthorrefmark{1}
SnT, University of Luxembourg, Luxembourg\\
chaima.medjadji@uni.lu, imane.hocine@uni.lu
}

\IEEEauthorblockA{\IEEEauthorrefmark{2}
FSTM/DCS, University of Luxembourg, Luxembourg\\
gregoire.danoy@uni.lu, sylvain.kubler@uni.lu, yves.letraon@uni.lu
}

\thanks{*Co-first authors.}
}

\maketitle

\begin{abstract}
Federated Learning (FL) enables collaborative training across decentralized clients, but most methods assume aligned feature schemas, an assumption that rarely holds in tabular settings where clients observe only partially overlapping feature subsets. In these heterogeneous feature spaces, parameter-averaging methods (e.g., FedAvg) transfer little information across weakly overlapping or disjoint feature groups, limiting their effectiveness for federated imputation. To overcome this, we propose \textbf{FedHF-Impute}, a federated imputation framework that separates structural feature unavailability from conventional missingness and uses a shared global feature graph to propagate information across statistically related features through message passing.  This enables indirect cross-client knowledge transfer—even when features are never jointly observed locally—while preserving standard federated communication. Under simulated partial schema overlap on the SECOM and AirQuality datasets, FedHF-Impute improves imputation accuracy (RMSE) over FL baselines by 26.9\%, and 8.4\% respectively, while achieving comparable performance on PhysioNET, with only a 0.3\% difference relative to the best baseline.

\end{abstract}


\section{Introduction}
Federated learning (FL) has emerged as a powerful paradigm for collaboratively training machine learning (ML) models across distributed clients while preserving data locality and privacy. Classical horizontal FL assumes that all clients share an identical feature space, enabling direct parameter aggregation, such as FedAvg. However, recent surveys on heterogeneous FL \cite{ye2023hflsurvey,lin2024towards,karami2025harmony} indicate that this assumption is frequently violated in practice, as real systems often differ in measurement pipelines, sensor configurations, and institutional data schemas. In such scenarios, certain features are \emph{structurally unavailable} on some clients. This violates the aligned-schema assumption foundational to most FL algorithms and leads to ill-posed aggregation when feature indices no longer correspond semantically. Recent work such as FLIC~\cite{rakotomamonjy2023flic} demonstrates that heterogeneous feature representations require explicit alignment or transformation even for standard predictive tasks. More recent studies~\cite{karami2025harmony,yu2025fedta} show that heterogeneity in feature spaces, data distributions, or temporal patterns can severely degrade both optimization stability and global model performance.
This motivates a clear separation between \emph{temporary missingness} \cite{mitra2023nature} and \emph{structural unavailability} \cite{nguyen2024fedmac}. Temporary missingness occurs when a feature is observable locally but has missing entries, whereas structural unavailability occurs when a client never measures that feature at all. Existing federated imputation methods primarily address temporary missingness and implicitly assume aligned schemas, limiting their applicability to realistic heterogeneous environments.



We address federated imputation by decoupling global feature semantics from local client schemas. Rather than tying features to local index positions, we model them as shared global entities linked by a global feature graph that captures statistical dependencies. This enables message passing and cross-client knowledge transfer, even between features never jointly observed on any client. Building on this idea, we introduce \textbf{FedHF-Impute}, a federated imputation framework that combines global feature-graph reasoning with standard FL optimization. The graph is built once from distributed statistics sent by the clients, then shared with all of them without adding training-time communication overhead. Clients perform local imputation using graph-informed feature embeddings, while the server aggregates parameters with FedAvg. FedHF-Impute distinguishes structural feature unavailability from missingness, supports indirect estimation of never co-observed feature interactions, and remains fully compatible with standard federated communication protocols.

\begin{table*}[b]
\centering
\begin{tabular}{llccc}
\toprule
Paradigm & Method & Schema Heterogeneity Support & Feature-Dependency Modeling & Graph-Based Propagation \\
\midrule
\multirow{5}{*}{Centralized} 
& MICE~\cite{buuren2011mice}                & -- & \cmark & \xmark \\
& DAE~\cite{vincent2008dae}                 & -- & \cmark & \xmark \\
& Transformer~\cite{vaswani2017attention}   & -- & \cmark & \xmark \\
& GAIN~\cite{yoon2018gain}                  & -- & \cmark & \xmark \\
& NewImp~\cite{newimp2024}                  & -- & \cmark & \xmark \\
\midrule
\multirow{8}{*}{Federated} 
& Fed-Mean~\cite{mcmahan2017fedavg}                       & \xmark & \xmark & \xmark \\

& Fed-DAE~\cite{mcmahan2017fedavg,vincent2008dae}         & \xmark & \cmark & \xmark \\
& Fed-Transformer~\cite{mcmahan2017fedavg,vaswani2017attention} & \xmark & \cmark & \xmark \\
& ED-FedImpute~\cite{miccai2025_edfedimpute}               & \xmark & \cmark & \xmark \\

& VFL-KNN-Impute~\cite{du2024vflknn}                       & \cmark & \xmark & \xmark \\
& FedFeatGen~\cite{poudel2025fedfeatgenmiua}               & \cmark & \cmark & \xmark \\
& \textbf{FedHF-Impute (ours)}                                    & \cmark & \cmark & \cmark \\
\bottomrule
\end{tabular}

\caption{Comparison of representative centralized and federated imputation methods.
\textit{Schema Heterogeneity Support} denotes the ability to operate under partially overlapping feature spaces across clients (e.g., vertical FL or missing modalities).
\textit{Feature-Dependency Modeling} refers to capturing inter-feature (or cross-modality) relationships beyond independent per-feature estimation.
\textit{Graph-Based Propagation} indicates whether feature interactions are explicitly structured and propagated through a learned graph.
Green checkmarks (\cmark) indicate explicit support, red crosses (\xmark) denote absence, and -- indicates that schema heterogeneity is not applicable or cannot be meaningfully assessed for centralized methods.}
\label{tab:method_comparison}
\end{table*}

We evaluate FedHF-Impute across multiple tabular datasets with controlled schema heterogeneity and show that it consistently outperforms existing federated and centralized imputation baselines. These findings align with recent observations that feature-space heterogeneity significantly impairs traditional FL optimization~\cite{ye2023hflsurvey,karami2025harmony}, underscoring the importance of structured feature-level modeling in next-generation federated systems.

The remainder of this paper is organized as follows. Section \ref{sec:baselines} reviews related work on FL with heterogeneous feature spaces and existing imputation approaches. Section \ref{sec:fedhf} presents the proposed FedHF-Impute framework, including the feature-centric modeling formulation, construction of the global feature graph, and the federated training procedure. Section \ref{sec:exp-set} describes the experimental setup, datasets, and evaluation protocol used to assess performance under varying degrees of schema heterogeneity, while the experimental results, including comparisons with baseline methods are reported in Section \ref{sec:exp-results}. Conclusion follows in Section \ref{sec:conclusion}.

\section{Related work}
\label{sec:baselines}

Imputation in FL environments intersects classical statistical methods, deep generative models, distributed optimization under non-IID data, and emerging techniques designed to address heterogeneous feature spaces. This section reviews these areas and situates \textit{FedHF-Impute} within the broader landscape of federated and heterogeneous-feature learning.

Classical imputers such as mean imputation, $k$NN, and MICE~\cite{buuren2011mice} have long been used for tabular datasets but struggle to capture nonlinear or high-dimensional dependencies. Deep neural imputers address these limitations using self-supervised reconstruction (DAE~\cite{vincent2008dae}) or adversarial objectives (GAIN~\cite{yoon2018gain}). Variational models such as MIWAE~\cite{mattei2019miwae} support probabilistic imputation under missing-at-random assumptions, while time-series models like BRITS~\cite{cao2018brits}, SAITS~\cite{du2023saits}, and CSDI~\cite{tashiro2021csdi} capture temporal or score-based dependencies. Transformer-based tabular models (e.g., TabTransformer~\cite{huang2020tabtransformer}) further demonstrate robustness to irregular missingness. Graph-based imputers such as GRIN~\cite{cini2022grin}, IGRM~\cite{zhong2023igrm}, and EGG-GAE~\cite{telyatnikov2023egggae} introduce relational inductive bias but still assume a globally aligned schema.

Recent diffusion-based imputers further advance centralized reconstruction quality. \textit{NewImp} formulates imputation as a negative-entropy regularized Wasserstein gradient flow, improving diffusion-based inference of missing values in tabular data~\cite{newimp2024}. These centralized architectures can tolerate substantial missingness via masking, attention mechanisms, or reconstruction modules. Moreover, recent work in time-series forecasting shows that centralized attention models can operate with partially observed variable subsets, for example via variable-reconstruction flows~\cite{ijcai2024_trf} or impute-then-forecast pipelines~\cite{adc2024_tiformer}. However, these approaches remain largely signal-centric and do not exploit external semantic or operational metadata that could guide valid conditioning context. Addressing this limitation, Vision~\cite{hocine2026vision} introduces a metadata-governed paradigm that combines knowledge graphs, diffusion-based imputation, and LLM-based orchestration to improve the quality, governance, and explainability of time series imputation. Nevertheless, such models still assume a globally pooled and fully enumerated feature space; they do not address schema heterogeneity in the federated setting, where some features are structurally absent for entire clients and dependencies must be inferred without ever observing all variables jointly.

Beyond early federated imputers, several peer-reviewed methods address distinct forms of heterogeneity. In vertical FL, \textit{Privacy-Preserving VFL KNN Imputation} performs feature imputation across parties with disjoint attribute subsets, explicitly handling schema heterogeneity~\cite{du2024vflknn}. For multimodal FL, \textit{FedFeatGen} reconstructs low-dimensional bottleneck features of absent modalities to enable collaboration across heterogeneous clients~\cite{poudel2025fedfeatgenmiua}. In medical imaging, an \textit{encoder-decoder federated imputation} approach replaces noisy or degraded images with reconstructed versions prior to training, improving robustness under federated constraints~\cite{miccai2025_edfedimpute}. While these methods model cross-modality or cross-view dependencies, they do not introduce an explicit feature-graph capable of propagating information across never co-observed features.

\begin{figure*}[b]
    \centering
    \includegraphics[width=\linewidth]{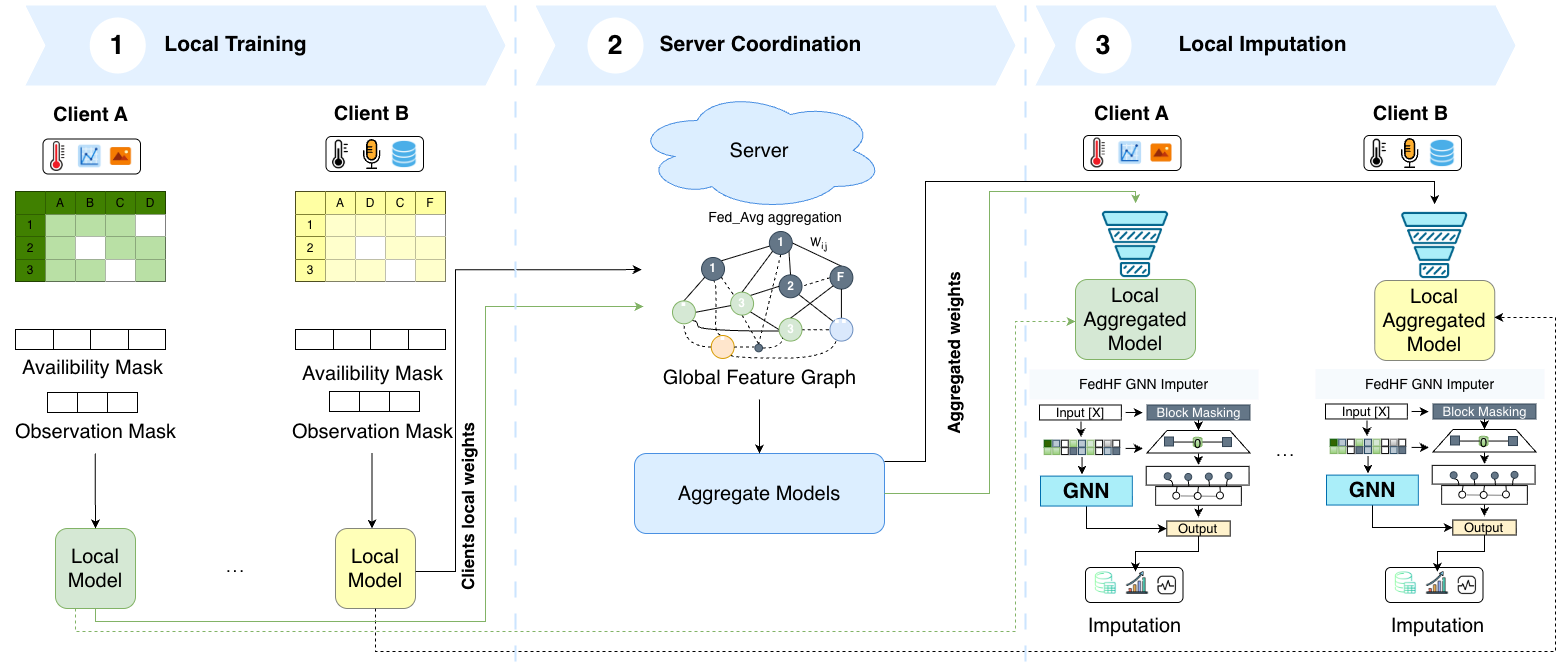}
    \caption{FedHF-Impute overview: Federated imputation with heterogeneous feature spaces}
    \label{ArchitectureFedhf}
\end{figure*}

Vertical Federated Learning (VFL) provides a natural setting for schema heterogeneity, where different organizations hold disjoint feature subsets for the same individuals. Surveys synthesize its architectures and privacy guarantees~\cite{yang2023vflsurvey,khan2025vflreview}. Entity alignment relies on secure matching protocols such as PSI~\cite{nist2024psi,nvflarepsi}. Methods such as SecureBoost~\cite{cheng2019secureboost}, SecureBoost+~\cite{fan2024secureboostplus}, and split learning~\cite{vepakomma2018split} further support vertical settings but assume perfectly matched entities and do not address partially overlapping schemas across autonomous clients.


Several federated imputation methods have been proposed. Fed-MIWAE~\cite{balelli2023fedmiwae} extends MIWAE to FL, and Cafe~\cite{min2024cafe} introduces personalized weighting for heterogeneous missingness. However, these works still assume globally aligned feature indices and therefore cannot model dependencies between features sparsely co-observed across clients. As Table~\ref{tab:method_comparison} shows, none employs a feature-graph capable of capturing cross-feature relations across heterogeneous schemas.

Recent work on heterogeneous feature spaces in FL seeks to support feature misalignment directly. FLIC~\cite{rakotomamonjy2023flic} aligns heterogeneous client features through optimal-transport barycenters, while FedSGCNN~\cite{suzuki2023fedsgcnn} clusters clients based on Siamese GNN similarity. More recent alignment approaches such as FedFA~\cite{zhou2024fedfa} and FedCOLA~\cite{chung2025fedcola} improve robustness under non-IID distributions but still assume shared feature indices. In contrast, \textit{FedHF-Impute} introduces a \emph{global feature graph} and performs message passing over feature nodes, explicitly distinguishing \emph{schema unavailability} from \emph{missingness}. This enables propagation of statistical strength between never co-observed features under strict data-locality constraints, a capability not addressed by existing federated or centralized models.

\section{FedHF-Impute: Federated Imputation under Heterogeneous Feature Spaces}
\label{sec:fedhf}

\subsection{Problem Formulation}

We consider a cross-silo federated learning setting comprising $K$ clients. Client $k$ maintains a local dataset $X^{(k)} \in \mathbb{R}^{N_k \times F}$, where $F$ denotes the \emph{global features} which are the union of all features that may appear across the federation.

In practice, clients exhibit \emph{schema heterogeneity}: each client collects measurements for only a subset of the globally defined features, reflecting differences in data collection protocols, sensor availability, or institutional constraints.

We characterize this heterogeneity through a two-level masking representation. The \emph{availability mask} $a^{(k)} \in \{0,1\}^{F}$ encodes the local schema of client $k$, where $a^{(k)}_f = 1$ if and only if feature $f$ is part of that client's data collection pipeline. 

Within the available schema, standard missing values are encoded by the \emph{observation mask} $m^{(k)} \in \{0,1\}^{N_k \times F}$, where $m^{(k)}_{i,f} = 1$ if $X^{(k)}_{i,f}$ is observed and $0$ otherwise. This distinction is fundamental: an entry with $a^{(k)}_f = 0$ is \emph{schema-unavailable} (the feature is structurally absent from the client's pipeline), whereas $a^{(k)}_f = 1$ with $m^{(k)}_{i,f} = 0$ constitutes a \emph{conventional missing value} within a collected feature. 

Standard imputation methods assume aligned feature spaces across all parties and therefore conflate these two conceptually distinct sources of absence while FedHF-Impute is specifically designed to address this gap.

\subsection{The FedHF-Impute Framework}

FedHF-Impute learns a single global imputation model shared across all clients while explicitly accommodating heterogeneous schemas. The method comprises three tightly coupled components: (i) a global feature relationship graph that captures inter-feature statistical dependencies, (ii) a GNN-based imputer that operates over the feature dimension, and (iii) a self-supervised block-masking objective that enables training without access to ground-truth missing values. 

All the details are explained in ALGORITHM~\ref{alg:fedhf-server} and \ref{alg:fedhf-client}. Now, we describe each component in turn:

\begin{algorithm*}[hbt!]
\caption{FedHF-Impute (Server)}
\label{alg:fedhf-server}
\begin{algorithmic}[1]
\Require Clients $\{1,\dots,K\}$, rounds $R$, global graph $G=(V,E,w)$, initial parameters $\theta^{(0)}$
\For{$t=0,\dots,R-1$}
  \State Select participating clients $S_t$ (e.g., uniformly at random)
  \State Broadcast $(\theta^{(t)}, \text{config}_t)$ to all $k\in S_t$ \Comment{Graph $G$ is pre-shared or referenced by path}
  \State Receive updated parameters $\{\theta^{(t)}_k\}_{k\in S_t}$ and weights $\{n_k\}_{k\in S_t}$
  \State $\theta^{(t+1)} \gets \sum_{k\in S_t}\frac{n_k}{\sum_{j\in S_t}n_j}\,\theta^{(t)}_k$ \Comment{FedAvg}
  \State Optionally aggregate and log validation RMSE from clients
\EndFor
\State \Return $\theta^{(R)}$
\end{algorithmic}
\end{algorithm*}

\begin{algorithm*}[h]
\caption{FedHF-Impute (Client $k$)}
\label{alg:fedhf-client}
\begin{algorithmic}[1]
\Require Local data $X^{(k)}$, availability mask $a^{(k)}$, graph $G$, local epochs $E$, block fraction $\rho$
\State Upon receiving $(\theta, \text{config})$ from server
  \State Load parameters $\theta$ into model $f_{\theta}$
  \For{epoch $=1,\dots,E$}
    \For{mini-batch $B$ from local training set}
      \State $x \gets \text{NaNs-to-0}(B)$; $m \gets \mathbb{1}[\text{observed}]$; $a \gets a^{(k)}$
      \State sample block feature set $F_{\mathrm{mask}}$ with $|F_{\mathrm{mask}}|\approx \rho|F_{\mathrm{avail}}|$
      \State $\Omega \gets \{(b,f): m_{b,f}=1 \wedge a_f=1 \wedge f\in F_{\mathrm{mask}}\}$
      \State Create corrupted inputs $(x',m')$ by setting $x'_{b,f}\!=\!0$ and $m'_{b,f}\!=\!0$ for $(b,f)\in\Omega$
      \State $\hat{x} \gets f_{\theta}(x', m', a, G)$
      \State $\mathcal{L} \gets \frac{1}{|\Omega|}\sum_{(b,f)\in\Omega}(\hat{x}_{b,f}-x_{b,f})^2$
      \State Update $\theta$ by one optimizer step minimizing $\mathcal{L}$
    \EndFor
  \EndFor
  \State Compute validation RMSE using deterministic $\Omega_{\mathrm{val}}$
  \State \Return updated parameters $\theta_k$ and metrics to the server.
\end{algorithmic}
\end{algorithm*}

\paragraph{Global feature graph}
The server first collects feature metadata from all clients (specifically, the list of feature identifiers each client holds, with the correlations between them computed on the training data using Pearson correlations) in order to establish the global feature set  $V = \{1, \dots, F\}$. Then, we construct a weighted feature graph $G = (V, E, w)$ with one node per feature in $V = \{1, \dots, F\}$. First, edge weights ($w_{ji}$) are derived from the correlations sent by clients. For each feature $i$, we retain edges to its top-$k$ ($k$ is a parameter of our model and it can be defined by the user) most correlated neighbors $j$ with weights
\[
  w_{ji} \;\in\; [0,1] \;\;\propto\;\; |\mathrm{corr}(i,j)|.
\]
The server then distributes the resulting edge structure $(\texttt{edge\_index},\,\texttt{edge\_weight})$ to all clients once, prior to federated training, and clients reuse this fixed graph throughout all communication rounds. This shared structure enables information propagation between correlated features even when a given client observes only a sparse subset of them.

\paragraph{GNN imputer architecture}
The imputer $f_{\theta}$ operates over the feature dimension rather than the sample dimension. Given a mini-batch of $B$ samples, we construct an input matrix $x \in \mathbb{R}^{B \times F}$ by replacing all missing entries with $0$, and supply two auxiliary tensors: the observation mask $m \in \{0,1\}^{B \times F}$ and the availability mask $a \in \{0,1\}^{F}$, broadcast across the batch. The model produces $\hat{x} = f_{\theta}(x, m, a, G) \in \mathbb{R}^{B \times F}$, providing predictions for every feature position.

Each feature $f$ is associated with a learnable embedding $e_f \in \mathbb{R}^{d}$. The initial node representation for sample $b$ and feature $f$ is computed by concatenating the feature embedding with the scalar observation, observation mask, and availability indicator, and projecting through a shared input MLP:
\begin{equation}
  h^{(0)}_{b,f}
  \;=\;
  \mathrm{MLP}_{\mathrm{in}}\!\Big(
    \bigl[\,e_f \;\Vert\; x_{b,f} \;\Vert\; m_{b,f} \;\Vert\; a_f\,\bigr]
  \Big)
  \;\in\; \mathbb{R}^{d}.
  \label{eq:input-enc}
\end{equation}
We then apply $L$ message-passing layers over $G$. At layer $\ell$, aggregated neighbor
signals and the current node state are combined via a residual update:
\begin{align}
  \mathrm{msg}^{(\ell)}_{b,i}
    &= \sum_{(j \to i) \in E} w_{ji}\, h^{(\ell)}_{b,j},
    \label{eq:aggregation}\\[4pt]
  h^{(\ell+1)}_{b,i}
    &= h^{(\ell)}_{b,i}
       + \mathrm{MLP}_{\ell}\!\Big(
           \bigl[\,h^{(\ell)}_{b,i} \;\Vert\; \mathrm{msg}^{(\ell)}_{b,i}\,\bigr]
         \Big).
  \label{eq:update}
\end{align}
A final output MLP maps each feature's last-layer representation to a scalar prediction:
\begin{equation}
  \hat{x}_{b,f}
  \;=\;
  \mathrm{MLP}_{\mathrm{out}}\!\bigl(h^{(L)}_{b,f}\bigr).
  \label{eq:pred-head}
\end{equation}

\paragraph{Self-supervised training objective}
Because ground-truth values for genuinely missing entries are unobserved, we rely on a self-supervised reconstruction objective. For each mini-batch, the client samples a \emph{corruption set} $\Omega$ drawn exclusively from entries that are both observed and schema-available:
\begin{equation}
  \Omega \;\subseteq\; \bigl\{(b, f) : m_{b,f} = 1 \;\wedge\; a_f = 1\bigr\}.
\end{equation}
We adopt \emph{block masking}: a subset of available features $F_{\mathrm{mask}} \subset F_{\mathrm{avail}}$ of cardinality $\lfloor\rho\,|F_{\mathrm{avail}}|\rfloor$ is drawn uniformly at random, and every eligible entry in those features is masked across the entire mini-batch. Corrupted inputs $(x', m')$ are formed by setting $x'_{b,f} = 0$ and $m'_{b,f} = 0$ for all $(b,f) \in \Omega$. The training objective is a mean-squared reconstruction loss evaluated only at corrupted positions:
\begin{equation}
  \mathcal{L}(\theta)
  \;=\;
  \frac{1}{|\Omega|}
  \sum_{(b,f)\,\in\,\Omega}
  \!\bigl(\hat{x}_{b,f} - x_{b,f}\bigr)^{2}.
  \label{eq:loss}
\end{equation}
At inference time, observed values are retained exactly and the model fills in only missing-but-available entries:
\begin{equation}
  X^{\mathrm{imp}}_{i,f}
  \;=\;
  \begin{cases}
    X_{i,f},       & m_{i,f} = 1, \\[2pt]
    \hat{x}_{i,f}, & m_{i,f} = 0 \;\wedge\; a_f = 1.
  \end{cases}
  \label{eq:inference}
\end{equation}
Schema-unavailable features ($a_f = 0$) are not imputed at any client where they are absent.

\subsection{Federated Training Protocol}

The imputer is trained via FedAvg over $R$ communication rounds. At round $t$, the server broadcasts the current global parameters $\theta^{(t)}$ to a selected cohort $S_t \subseteq [K]$. Each client runs $E$ local epochs of stochastic gradient descent on its private dataset and returns the updated parameters $\theta^{(t)}_k$. The server aggregates these updates proportionally to local dataset sizes:
\begin{equation}
  \theta^{(t+1)}
  \;=\;
  \sum_{k \in S_t}
  \frac{n_k}{\sum_{j \in S_t} n_j}\,\theta^{(t)}_k,
  \label{eq:fedavg}
\end{equation}
where $n_k$ denotes the number of training samples held by client $k$. Throughout training, only model parameters and scalar validation metrics are communicated to the server while raw data never leave the client, preserving the privacy guarantees inherent to the federated setting.

For convergence monitoring, each client evaluates validation RMSE on a held-out corruption mask $\Omega_{\mathrm{val}}$ generated with a fixed random seed and reports the resulting scalar to the server for aggregation and logging.

\section{Experimental Setup}
\label{sec:exp-set}
To evaluate the effectiveness of the proposed approach, we conduct a series of experiments. This section describes the experimental setup, including the datasets, the federated learning settings, the implementation details, and the evaluation protocol.
\paragraph{Datasets}
We conduct experiments on three real-world tabular datasets exhibiting distinct structural characteristics and dimensionality. 

\textbf{AirQuality}\footnote{https://archive.ics.uci.edu/dataset/360/air+quality} consists of environmental sensor measurements with moderate feature dimensionality and correlated pollutant variables (13 features). 

\textbf{PhysioNET}\footnote{https://datasets.bio/source/PhysioNet} is a clinical time-series dataset containing physiological measurements from intensive care patients. After preprocessing, it forms a medium-dimensional tabular representation with clinically correlated variables and naturally occurring missingness (42 features).

\textbf{SECOM}\footnote{https://archive.ics.uci.edu/dataset/179/secom} is a high-dimensional industrial manufacturing dataset characterized by substantial missingness, noisy measurements, and strong inter-feature correlations across hundreds of sensors (590 features).

To emulate realistic heterogeneous feature environments, we simulate partial schema overlap by assigning each client a random 60\% subset of features kept from the dataset.  This is implemented via a binary feature-availability mask, ensuring that certain features are structurally unavailable on specific clients. This setup reflects practical federated scenarios where institutions collect different but partially overlapping measurements.

\paragraph{Federated Settings}
The data are partitioned across $K=4$ clients, each maintaining local training data that contain naturally occurring missing values. Prior to federated training, clients send feature metadata (the list of feature identifiers they hold) to the server, which uses this information to construct the global feature graph. No raw data is shared in this step. Communication during training occurs exclusively through model parameter exchange coordinated by the central server. A global test set, constructed from the union of all features, is used strictly for evaluation and is never accessible during training.

\paragraph{Implementation Details}
All features are standardized prior to training to ensure consistent scaling across datasets. Neural baselines are optimized using the Adam optimizer, with early stopping based on validation performance. In the federated setting, clients perform multiple local gradient updates per communication round before server aggregation using FedAvg for baseline methods. FedHF-Impute follows the same communication schedule and data partitions to ensure a fair comparison. All reported results are obtained using identical data splits, feature-availability masks, and corruption patterns across methods. Our source code is publicly available.\footnote{ https://github.com/chaimaamedjadji/FedHFImpute}

\paragraph{Evaluation Protocol}
To evaluate imputation quality in a controlled and comparable manner, we introduce additional artificial missingness on the test set by randomly masking a fixed proportion of entries that are both originally observed and structurally available. Performance is measured using \textbf{Root Mean Squared Error (RMSE)} computed exclusively on these artificially corrupted positions, ensuring that evaluation reflects true reconstruction accuracy.

\paragraph{Model selection and uncertainty reporting.}
To ensure a fair comparison, all neural federated methods use the same communication budget, client partitions, feature-availability masks, and artificial test corruption masks. Model selection is performed using validation RMSE only; final reported RMSE is computed on the held-out test corruption mask, which is never used for training or checkpoint selection. Centralized baselines are trained separately on each client-specific schema and therefore report mean $\pm$ standard deviation across clients. Federated methods are trained collaboratively and evaluated on the common federated test protocol. We additionally report variability across repeated runs whenever available. This distinction reflects the different evaluation units of centralized local imputers and federated global imputers, but all methods are scored on identical corrupted entries.

For centralized baselines, we report the \emph{RMSE} and \emph{standard deviation} of RMSE across clients to capture variability induced by heterogeneous feature subsets. For federated methods, we report the best validation RMSE across communication rounds, reflecting convergence during collaborative training.

\section{Results \& Discussion}
\label{sec:exp-results}

\begin{table*}[hbt!]
\centering
\caption{Imputation performance on AirQuality, PhysioNET and SECOM (RMSE $\downarrow$). 
All methods are evaluated on the same held-out artificial corruption masks. 
For centralized baselines, mean $\pm$ std is computed across client-specific schema subsets. 
For federated methods, RMSE is computed from the selected validation checkpoint and evaluated on the common held-out test corruption mask. 
Best results are highlighted in bold within each learning paradigm.}
\label{tab:unified_grouped_results}
\resizebox{\linewidth}{!}{%
\begin{tabular}{llccc}
\toprule
Paradigm & Method & AirQuality & PhysioNET & SECOM \\
\midrule

\multirow{4}{*}{\textbf{Centralized}}
& MICE         & \textbf{$0.6756 \pm 0.0552$} & $1.1821 \pm 0.4247$ & $1.1231 \pm 0.1425$ \\
& DAE          & $0.6821 \pm 0.0261$ & \textbf{$0.9500 \pm 0.0268$} & $20.3387 \pm 35.5735$ \\
& Transformer  & $0.7080 \pm 0.0382$ & $0.9824 \pm 0.0186$ & \textbf{$1.1381 \pm 0.1343$} \\
& GAIN         & $0.9866 \pm 0.0499$ & $1.0295 \pm 0.0240$ & $27.5622 \pm 48.6025$ \\
\midrule

\multirow{6}{*}{\textbf{Federated}}
& Fed-Mean         & $0.9902 \pm 0.0079$ & $0.9927 \pm 0.0194$ & $1.1762 \pm 0.1529$ \\
& FED-DAE          & $0.7726 \pm 0.0356$ & $0.9718 \pm 0.0228$ & $4.0519 \pm 2.8018$ \\
& Fed-Transformer  & $0.9931 \pm 0.0319$ & $1.0367 \pm 0.0124$ & $1.2300 \pm 0.1497$ \\
& VFL-KNN-Impute   & $0.9889 \pm 0.0101$ & $0.9662 \pm 0.0329$ & $1.1231 \pm 0.1425$ \\
& Fed-FeatGen      & $0.8233 \pm 0.0249$ & {\boldmath\bfseries$0.9339 \pm 0.0181$} & $2.2202 \pm 1.1321$ \\
& \textbf{FedHF-Impute} & {\boldmath\bfseries$0.7074 \pm 0.0729$} & $0.9363 \pm 0.0367$ & {\boldmath\bfseries$0.8205 \pm 0.0791$} \\
\bottomrule
\end{tabular}%
}
\end{table*}

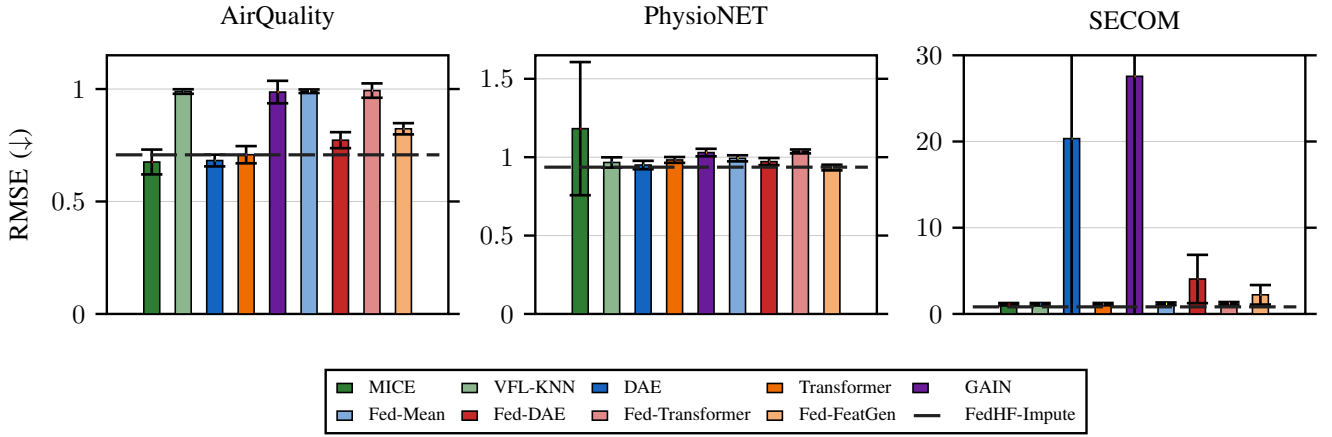
\begin{figure*}[t]
\centering
\begin{tikzpicture}

\definecolor{Cgreen}{HTML}{2E7D32}
\definecolor{Cblue}{HTML}{1565C0}
\definecolor{Cred}{HTML}{C62828}
\definecolor{Corange}{HTML}{EF6C00}
\definecolor{Cblack}{HTML}{212121}
\definecolor{Cpurple}{HTML}{6A1B9A}

\colorlet{CgreenL}{Cgreen!55}
\colorlet{CblueL}{Cblue!55}
\colorlet{CredL}{Cred!55}
\colorlet{CorangeL}{Corange!55}

\def\xA{-1.40}
\def\xB{-1.05}
\def\xC{-0.70}
\def\xD{-0.35}
\def\xE{ 0.00}
\def\xF{ 0.35}
\def\xG{ 0.70}
\def\xH{ 1.05}
\def\xI{ 1.40}

\pgfplotsset{
  cycle list/Set1-9,
  myaxis/.style={
    axis lines=box,
    line width=0.9pt,
    tick style={line width=0.9pt, black},
    tick align=outside,
    ymajorgrids=true,
    grid style={black!20},
    xmin=-1.9, xmax=1.9,
    xtick=\empty,
    ylabel={RMSE ($\downarrow$)},
  },
  mybar/.style={
    ybar,
    bar width=6pt,
    draw=black,
    line width=0.6pt,
  },
  whisker/.style={
    only marks,
    mark=-,
    mark size=0pt,
    error bars/y dir=both,
    error bars/y explicit,
    /pgfplots/error bars/error bar style={black, line width=1.0pt},
    /pgfplots/error bars/error mark options={black, line width=1.0pt, mark size=4pt, rotate=90},
  },
}

\begin{groupplot}[
  group style={group size=3 by 1, horizontal sep=1.15cm},
  width=6.1cm,
  height=5.0cm,
]

\nextgroupplot[
  myaxis,
  title={AirQuality},
  ymin=0, ymax=1.15,
  legend to name=SharedLegend,
  legend style={
    draw=black, fill=white, line width=0.9pt,
    legend columns=5,
    font=\scriptsize,
    /tikz/column sep=6pt,
  },
]

\addplot+[mybar, fill=Cgreen, area legend]   coordinates {(\xA,0.6756)}; \addlegendentry{MICE}
\addplot+[mybar, fill=CgreenL, area legend]  coordinates {(\xB,0.9889)}; \addlegendentry{VFL-KNN}
\addplot+[mybar, fill=Cblue, area legend]    coordinates {(\xC,0.6821)}; \addlegendentry{DAE}
\addplot+[mybar, fill=Corange, area legend]  coordinates {(\xD,0.7080)}; \addlegendentry{Transformer}
\addplot+[mybar, fill=Cpurple, area legend]  coordinates {(\xE,0.9866)}; \addlegendentry{GAIN}
\addplot+[mybar, fill=CblueL, area legend]   coordinates {(\xF,0.9902)}; \addlegendentry{Fed-Mean}
\addplot+[mybar, fill=Cred, area legend]     coordinates {(\xG,0.7726)}; \addlegendentry{Fed-DAE}
\addplot+[mybar, fill=CredL, area legend]    coordinates {(\xH,0.9931)}; \addlegendentry{Fed-Transformer}
\addplot+[mybar, fill=CorangeL, area legend] coordinates {(\xI,0.8233)}; \addlegendentry{Fed-FeatGen}

\addplot+[whisker] coordinates {(\xA,0.6756) +- (0,0.0552)};
\addplot+[whisker] coordinates {(\xB,0.9889) +- (0,0.0101)};
\addplot+[whisker] coordinates {(\xC,0.6821) +- (0,0.0261)};
\addplot+[whisker] coordinates {(\xD,0.7080) +- (0,0.0382)};
\addplot+[whisker] coordinates {(\xE,0.9866) +- (0,0.0499)};
\addplot+[whisker] coordinates {(\xF,0.9902) +- (0,0.0079)};
\addplot+[whisker] coordinates {(\xG,0.7726) +- (0,0.0356)};
\addplot+[whisker] coordinates {(\xH,0.9931) +- (0,0.0319)};
\addplot+[whisker] coordinates {(\xI,0.8233) +- (0,0.0249)};

\addplot+[very thick, dash pattern=on 10pt off 3pt, Cblack, forget plot]
coordinates {(-1.8,0.7074) (1.8,0.7074)};

\addlegendimage{
  legend image code/.code={
    \draw[very thick, dash pattern=on 10pt off 3pt, Cblack] (0cm,0cm) -- (0.45cm,0cm);
  }
}
\addlegendentry{FedHF-Impute}

\nextgroupplot[
  myaxis,
  title={PhysioNET},
  ymin=0, ymax=1.65,
  ylabel={},
]

\addplot+[mybar, fill=Cgreen]   coordinates {(\xA,1.1821)};
\addplot+[mybar, fill=CgreenL]  coordinates {(\xB,0.9662)};
\addplot+[mybar, fill=Cblue]    coordinates {(\xC,0.9500)};
\addplot+[mybar, fill=Corange]  coordinates {(\xD,0.9824)};
\addplot+[mybar, fill=Cpurple]  coordinates {(\xE,1.0295)};
\addplot+[mybar, fill=CblueL]   coordinates {(\xF,0.9927)};
\addplot+[mybar, fill=Cred]     coordinates {(\xG,0.9718)};
\addplot+[mybar, fill=CredL]    coordinates {(\xH,1.0367)};
\addplot+[mybar, fill=CorangeL] coordinates {(\xI,0.9339)};

\addplot+[whisker] coordinates {(\xA,1.1821) +- (0,0.4247)};
\addplot+[whisker] coordinates {(\xB,0.9662) +- (0,0.0329)};
\addplot+[whisker] coordinates {(\xC,0.9500) +- (0,0.0268)};
\addplot+[whisker] coordinates {(\xD,0.9824) +- (0,0.0186)};
\addplot+[whisker] coordinates {(\xE,1.0295) +- (0,0.0240)};
\addplot+[whisker] coordinates {(\xF,0.9927) +- (0,0.0194)};
\addplot+[whisker] coordinates {(\xG,0.9718) +- (0,0.0228)};
\addplot+[whisker] coordinates {(\xH,1.0367) +- (0,0.0124)};
\addplot+[whisker] coordinates {(\xI,0.9339) +- (0,0.0181)};

\addplot+[very thick, dash pattern=on 10pt off 3pt, Cblack, forget plot]
coordinates {(-1.8,0.9363) (1.8,0.9363)};

\nextgroupplot[
  myaxis,
  title={SECOM},
  ymin=0, ymax=30,
  ylabel={},
]

\addplot+[mybar, fill=Cgreen]   coordinates {(\xA,1.1231)};
\addplot+[mybar, fill=CgreenL]  coordinates {(\xB,1.1231)};
\addplot+[mybar, fill=Cblue]    coordinates {(\xC,20.3387)};
\addplot+[mybar, fill=Corange]  coordinates {(\xD,1.1381)};
\addplot+[mybar, fill=Cpurple]  coordinates {(\xE,27.5622)};
\addplot+[mybar, fill=CblueL]   coordinates {(\xF,1.1762)};
\addplot+[mybar, fill=Cred]     coordinates {(\xG,4.0519)};
\addplot+[mybar, fill=CredL]    coordinates {(\xH,1.2300)};
\addplot+[mybar, fill=CorangeL] coordinates {(\xI,2.2202)};

\addplot+[whisker] coordinates {(\xA,1.1231) +- (0,0.1425)};
\addplot+[whisker] coordinates {(\xB,1.1231) +- (0,0.1425)};
\addplot+[whisker] coordinates {(\xC,20.3387) +- (0,35.5735)};
\addplot+[whisker] coordinates {(\xD,1.1381) +- (0,0.1343)};
\addplot+[whisker] coordinates {(\xE,27.5622) +- (0,48.6025)};
\addplot+[whisker] coordinates {(\xF,1.1762) +- (0,0.1529)};
\addplot+[whisker] coordinates {(\xG,4.0519) +- (0,2.8018)};
\addplot+[whisker] coordinates {(\xH,1.2300) +- (0,0.1497)};
\addplot+[whisker] coordinates {(\xI,2.2202) +- (0,1.1321)};

\addplot+[very thick, dash pattern=on 10pt off 3pt, Cblack, forget plot]
coordinates {(-1.8,0.8205) (1.8,0.8205)};

\end{groupplot}

\node[anchor=north] at ($(group c1r1.south)!0.5!(group c3r1.south) + (0,-0.60cm)$) {%
\begin{tikzpicture}[baseline]
\matrix[
  matrix of nodes,
  nodes={font=\scriptsize, anchor=west, text depth=0.2ex, text height=1.0ex},
  column sep=2pt,
  row sep=1pt,
  inner sep=2pt,
  draw=black,
  line width=0.9pt,
  fill=white,
] {
  \tikz{\draw[draw=black, fill=Cgreen,  line width=0.6pt] (0,0) rectangle (0.20,0.12);} & MICE &
  \tikz{\draw[draw=black, fill=CgreenL, line width=0.6pt] (0,0) rectangle (0.20,0.12);} & VFL-KNN &
  \tikz{\draw[draw=black, fill=Cblue,   line width=0.6pt] (0,0) rectangle (0.20,0.12);} & DAE &
  \tikz{\draw[draw=black, fill=Corange, line width=0.6pt] (0,0) rectangle (0.20,0.12);} & Transformer &
  \tikz{\draw[draw=black, fill=Cpurple, line width=0.6pt] (0,0) rectangle (0.20,0.12);} & GAIN \\
  \tikz{\draw[draw=black, fill=CblueL,  line width=0.6pt] (0,0) rectangle (0.20,0.12);} & Fed-Mean &
  \tikz{\draw[draw=black, fill=Cred,    line width=0.6pt] (0,0) rectangle (0.20,0.12);} & Fed-DAE &
  \tikz{\draw[draw=black, fill=CredL,   line width=0.6pt] (0,0) rectangle (0.20,0.12);} & Fed-Transformer &
  \tikz{\draw[draw=black, fill=CorangeL,line width=0.6pt] (0,0) rectangle (0.20,0.12);} & Fed-FeatGen &
  \tikz{\draw[very thick, dash pattern=on 10pt off 3pt, Cblack] (0,0.06) -- (0.45,0.06);} & FedHF-Impute \\
};
\end{tikzpicture}%
};

\end{tikzpicture}
\caption{Imputation performance (RMSE $\downarrow$) of FedHF-Impute vs the other baselines on three datasets: AirQuality, PhysioNET and SECOM.}
\label{fig:fedhf_baselines}
\end{figure*}

We evaluate FedHF-Impute against centralized (MICE, DAE, Transformer, GAIN) and federated 
(Fed-Mean, Fed-DAE, Fed-Transformer, VFL-KNN, Fed-FeatGen) baselines under heterogeneous 
feature settings on AirQuality, PhysioNET, and SECOM. All results are reported as 
standardized RMSE (z-space), summarized in Table~\ref{tab:unified_grouped_results} and 
Figure~\ref{fig:fedhf_baselines}.

The central finding is that FedHF-Impute is the only method that consistently performs 
well across all three datasets, and that its advantage grows with feature dimensionality 
and schema heterogeneity. Crucially, it achieves this while preserving privacy and 
operating without any centralized data access.

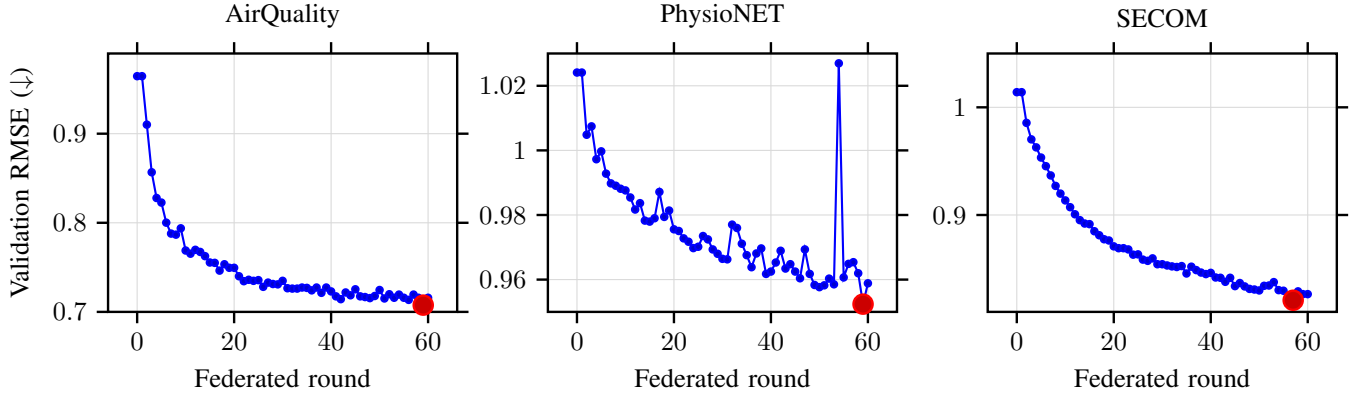
\begin{figure*}[hbt!]
\centering
\begin{tikzpicture}

\pgfplotsset{
  myaxis/.style={
    axis lines=box,
    line width=0.9pt,
    tick style={line width=0.9pt, black},
    tick align=outside,
    grid=both,
    grid style={black!15},
    xlabel={Federated round},
    ylabel={Validation RMSE ($\downarrow$)},
  }
}

\begin{groupplot}[
  group style={group size=3 by 1, horizontal sep=1.2cm},
  width=6.2cm,
  height=5.0cm,
  myaxis,
  legend to name=SharedLegend,
  legend style={
    draw=black, fill=white, line width=0.9pt,
    font=\scriptsize,
    legend columns=2,
    /tikz/column sep=10pt,
  },
]

\nextgroupplot[
  title={AirQuality},
  ymin=0.70, ymax=0.99,
]
\addplot+[thick, mark=*, mark size=1.2pt]
table[col sep=comma, x=round, y=val_rmse]{results/fedhf_airquality_digitized.csv};
\addlegendentry{FedHF-Impute}
\addplot+[only marks, mark=*, mark size=3.6pt]
coordinates {(59,0.7076)};
\addlegendentry{Best round}

\nextgroupplot[
  title={PhysioNET},
  ymin=0.95, ymax=1.03,
  ylabel={},
]
\addplot+[thick, mark=*, mark size=1.2pt]
table[col sep=comma, x=round, y=val_rmse]{results/fedhf_physionet_digitized.csv};
\addplot+[only marks, mark=*, mark size=3.6pt]
coordinates {(59,0.9524)};

\nextgroupplot[
  title={SECOM},
  ymin=0.81, ymax=1.05,
  ylabel={},
]
\addplot+[thick, mark=*, mark size=1.2pt]
table[col sep=comma, x=round, y=val_rmse]{results/fedhf_secom_digitized.csv};
\addplot+[only marks, mark=*, mark size=3.6pt]
coordinates {(57,0.8208)};

\end{groupplot}

\node[anchor=north] at ($(group c1r1.south)!0.5!(group c3r1.south) + (0,-0.45cm)$)
{\ref{SharedLegend}};

\end{tikzpicture}
\caption{FedHF-Impute convergence across federated rounds on AirQuality, PhysioNET, and SECOM. Each curve shows the validation RMSE ($\downarrow$) at every communication round, and the highlighted marker denotes the best validation RMSE achieved during training for each dataset.}
\label{fig:fedhf_courves}
\end{figure*}

\subsection{Robustness to Schema Heterogeneity}

A primary objective of FedHF-Impute is to mitigate the performance degradation observed 
when neural imputers are trained under partially overlapping feature schemas via naive 
federated averaging.

Table~\ref{tab:unified_grouped_results} and ~\figurename~\ref{fig:fedhf_baselines} confirm that this degradation is systematic 
across all datasets and architectures. 

On AirQuality, DAE degrades from 0.6821 
(centralized) to 0.7726 (Fed-DAE), and Transformer-based imputation deteriorates from 0.7080 to 0.9931 under FedAvg. 

On PhysioNET, Transformer increases from 0.9824 to 
1.0367, and DAE from 0.9500 to 0.9718. More recent federated methods partially close this gap: VFL-KNN reaches 0.9662 on PhysioNET and Fed-FeatGen reaches 0.9339, yet  both still fall short of centralized DAE (0.9500). This confirms that parameter aggregation alone is insufficient when clients observe different feature subsets, and that vertical federated extensions offer only partial relief.

FedHF-Impute substantially reduces this degradation across all three datasets. 
On AirQuality, it achieves 0.7074, outperforming every federated baseline and closely 
approaching the strongest centralized method (MICE, 0.6756). On PhysioNET, it achieves 0.9363, outperforming all federated baselines except Fed-FeatGen (0.9339), which 
achieves a marginally lower RMSE on this dataset. On SECOM, the improvement is most striking: while federated baselines range from 1.1762 (Fed-Mean) to 4.0519 (Fed-DAE), FedHF-Impute attains 0.8205, outperforming every federated and centralized competitor 
by a wide margin.

These results demonstrate that feature-level graph modeling directly addresses the 
schema-mismatch failure mode of FedAvg, and that it generalizes across datasets of 
varying complexity.

\subsection{Competitiveness with Centralized Learning}

A federated method is only practically useful if it remains competitive with its 
centralized counterparts. The results in Table~\ref{tab:unified_grouped_results} and ~\figurename~\ref{fig:fedhf_baselines} show 
that FedHF-Impute achieves this across all three settings, and surpasses centralized 
baselines on the most challenging one.

On AirQuality, FedHF-Impute (0.7074) approaches MICE (0.6756), the best centralized 
method, while outperforming centralized DAE (0.6821) and Transformer (0.7080). On 
PhysioNET, the gap to the best centralized model (DAE, 0.9500) is small (0.9363 vs.\ 
0.9500) and the method outperforms centralized MICE and GAIN by a large margin.

On SECOM, FedHF-Impute achieves 0.8205 against the best centralized results of 1.1231 
(MICE) and 1.1381 (Transformer), a reduction of more than 27\%. This is particularly 
notable because generative (GAIN: 27.5622) and autoencoding (DAE: 20.3387) approaches 
suffer severe instability on this high-dimensional industrial dataset, indicating that 
independent per-feature modeling breaks down when feature correlations are numerous and 
locally distributed. Graph-based propagation, by contrast, explicitly structures these 
correlations, which explains why FedHF-Impute not only avoids this instability but 
improves upon all centralized alternatives.

\subsection{Scaling Behaviour with Feature Dimensionality}

The results suggest a consistent pattern showing that the advantage of FedHF-Impute grows with dataset dimensionality and structural complexity. In moderate-dimensional settings 
(AirQuality, PhysioNET), improvements over the best federated competitors are meaningful but modest. In the high-dimensional SECOM setting, where feature correlations are numerous and distributed across heterogeneous client schemas, the margin becomes substantial.

We note that this conclusion is based on three datasets spanning a range of 
dimensionalities, and we do not claim a formally established scaling law. However, 
the pattern is consistent with the intuition that graph-based feature propagation 
becomes increasingly advantageous as local per-feature estimation grows ill-conditioned 
dimensionality ablations would help to verify this hypothesis more rigorously.

\subsection{Optimization Stability}

~\figurename~\ref{fig:fedhf_courves} shows the evolution of FedHF-Impute validation RMSE 
across communication rounds. On AirQuality and PhysioNET, best validation RMSE is 
reached at round 59 (0.7074 and 0.9363, respectively), and on SECOM at round 57 
(0.8205). In all three cases, the validation curves decrease monotonically without 
oscillatory divergence, indicating that the federated optimization procedure is well-behaved.

This stability contrasts with the high cross-client RMSE exhibited by federated 
baselines (Fed-DAE on SECOM: std $\approx$ 2.80, Fed-FeatGen on SECOM: std $\approx$ 1.13), which reflects their sensitivity to the specific feature subset observed 
by each client.

After analyzing the results summarized in ~\tablename~\ref{tab:unified_grouped_results}, ~\figurename~\ref{fig:fedhf_baselines} and ~\ref{fig:fedhf_courves}, we find FedHF-Impute as a robust and stable approach to heterogeneous federated imputation, with advantages that scale with dataset complexity.

\subsection{Privacy of graph-construction statistics.}
Although FedHF-Impute does not transmit raw records, the feature graph is constructed from client-provided feature identifiers and pairwise correlation estimates. These statistics may reveal distributional information, especially in small cohorts or sensitive domains. In the present work, we treat graph construction as a pre-training metadata exchange step under the same trust assumptions commonly used in cross-silo FL. In stricter deployments, the graph-construction phase can be protected using secure aggregation, thresholding of low-support correlations, correlation clipping, or differentially private noise addition before server-side graph construction. We therefore view privacy-preserving graph construction as an important extension rather than a solved component of the current framework.

\subsection{Limitations.}
Our experiments use controlled schema heterogeneity with $K=4$ clients and random 60\% feature availability. This setting is appropriate for isolating the effect of partial feature overlap, but it does not cover all real cross-institutional deployment conditions. In particular, larger client cohorts, institution-specific schema patterns, non-random feature availability, and real multi-site datasets may introduce additional optimization and privacy challenges. Future work will therefore evaluate FedHF-Impute under larger client populations and naturally heterogeneous schemas.
\section{Conclusion}
\label{sec:conclusion}

In this work, we presented \textit{FedHF-Impute}, a federated imputation framework designed for the realistic but underexplored setting where clients hold only partially overlapping feature schemas. By constructing a global feature graph from client metadata and performing message passing over feature nodes, the method enables cross-client learning of 
inter-feature dependencies without sharing raw data or modifying the standard federated 
communication protocol.

Experiments on AirQuality, PhysioNET, and SECOM show that FedHF-Impute outperforms federated baselines on AirQuality and SECOM, achieves comparable performance on PhysioNET, and surpasses all centralized competitors on the high-dimensional SECOM benchmark (0.8205 vs.\ 1.1231 for the best centralized method), while converging stably across communication rounds. These results confirm that explicitly modeling feature-level structure is essential for effective 
imputation under schema heterogeneity.

We aim in future work to include learning adaptive graph structures during training, incorporating differential privacy for graph construction to protect feature metadata,
combining FedHF-Impute with communication-efficient federated optimization~\cite{chaimaa2025fedsparq} for reducing bandwidth overhead in large-scale deployments, extending message-passing to richer neighborhood functions, and broadening the framework to temporal and multimodal federated settings.

\section*{Acknowledgment}
This work was funded in whole or in part by the Luxembourg National Research Fund (FNR) under the LightGrid-SEED Project (ref. C21/IS/16215802/LightGridSEED) and supported by the Luxembourg National Research Fund (FNR) and the National Centre for Research and Development (NCBR) under the SERENITY Project (ref. C22/IS/17395419; POLLUX-XI/15/Serenity/2023).

\end{document}